\pdfoutput=1

\documentclass[11pt]{article}

\usepackage[]{acl}

\usepackage{times}
\usepackage{latexsym}

\usepackage[T1]{fontenc}

\usepackage[utf8]{inputenc}

\usepackage{microtype}

\usepackage{graphicx}
\usepackage{booktabs,tabularx,caption}

\usepackage{microtype}
\usepackage{graphicx}
\usepackage{multirow}

\usepackage{amsmath}
\usepackage{bbm}
\usepackage{tabularx}
\newcolumntype{Y}{>{\centering\arraybackslash}X}
\usepackage{booktabs}
\usepackage{algpseudocode, algorithmicx, algorithm}
\usepackage{enumitem}

\usepackage{hyperref}

\title{Program Transfer for Answering Complex Questions \\over Knowledge Bases}

  \newcommand*{\email}[1]{\texttt{#1}}

  \author{
    \textbf{Shulin Cao}$^{1,2}$, \textbf{Jiaxin Shi}$^{1,3}$, \textbf{Zijun Yao}$^{1,2}$, \textbf{Xin Lv}$^{1,2}$, \textbf{Jifan Yu}$^{1,2}$, \\
    \textbf{Lei Hou}$^{1,2}$\thanks{\quad Corresponding Author}\hspace{0.5em}, \textbf{Juanzi Li}$^{1,2}$, \textbf{Zhiyuan Liu}$^{1,2}$, \textbf{Jinghui Xiao}$^{4}$ \\
    $^1$Department of Computer Science and Technology, BNRist \\
    $^2$KIRC, Institute for Artificial Intelligence, Tsinghua University, Beijing 100084, China \\
    $^3$Cloud BU, Huawei Technologies,
    $^4$Noah's Ark Lab, Huawei Technologies\\
    \email{\{caosl19, yaozj20, lv-x18\}@mails.tsinghua.edu.cn}\\
    \email{shijx12@gmail.com},
    \email{\{houlei,lijuanzi\}tsinghua.edu.cn}\\
   \\
    }

\begin{document}
\maketitle
\begin{abstract}
Program induction for answering complex questions over knowledge bases (KBs) aims to decompose a question into a multi-step program, whose execution against the KB produces the final answer. 
Learning to induce programs relies on a large number of parallel question-program pairs for the given KB. 
However, for most KBs, the gold program annotations are usually lacking, making learning difficult.
In this paper, we propose the approach of \textbf{program transfer}, which aims to leverage the valuable program annotations on the rich-resourced KBs as external supervision signals to aid program induction for the low-resourced KBs  that lack program annotations. 
For program transfer, we design a novel two-stage parsing framework with an efficient ontology-guided pruning strategy. 
First, a sketch parser translates the question into a high-level program sketch, which is the composition of functions. 
Second, given the question and sketch, an argument parser searches the detailed arguments from the KB for functions. 
During the searching, we incorporate the KB ontology to prune the search space. 
The experiments on ComplexWebQuestions and WebQuestionSP show that our method outperforms SOTA methods significantly, demonstrating the effectiveness of program transfer and our framework.  Our codes and datasets
can be obtained from \url{https://github.com/THU-KEG/ProgramTransfer}.

\end{abstract}

\section{Introduction}

\begin{figure}[!t]
        \centering
        \includegraphics[width=1.0\columnwidth]{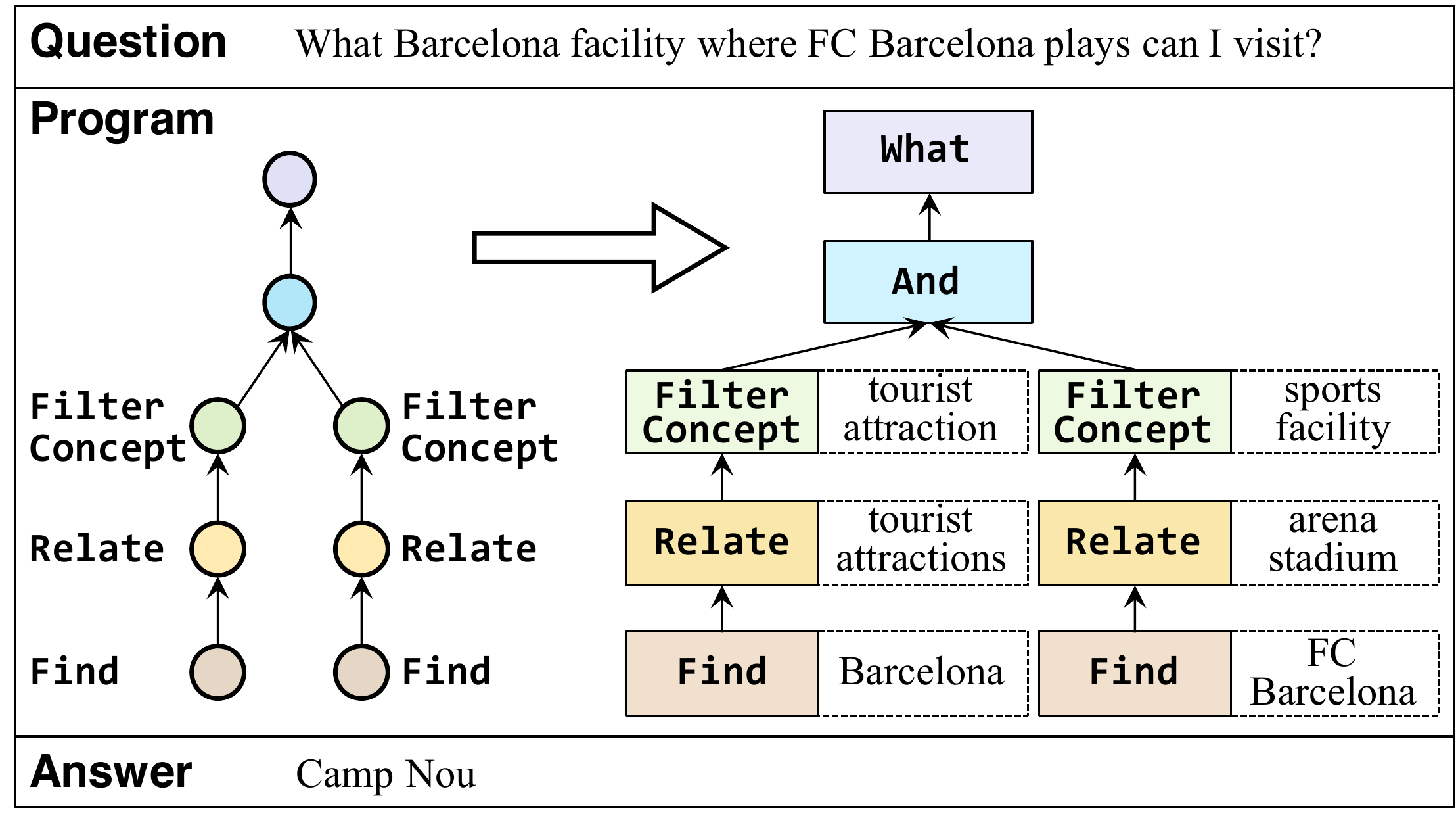} 
        \caption{An example question, the corresponding program, and the answer. The left side is the sketch, and the right side is the complete program, with dotted boxes denoting arguments for functions.}
        \label{fig:example}
\end{figure}

Answering complex questions over knowledge bases (Complex KBQA) is a challenging task requiring logical, quantitative, and comparative reasoning over KBs~\cite{Hu2018ASF,Lan2021ASO}. 
Recently, the program induction (PI) paradigm, which gains increasing study in various areas~\cite{Lake2015HumanlevelCL,Neelakantan2017LearningAN,Wong2021LeveragingLT}, emerges as a promising technique for Complex KBQA~\cite{NSM,Saha2019ComplexPI,NPI}. 
Given a KB, PI for Complex KBQA aims to decompose a complex question into a multi-step program, whose execution on the KB produces the answer.
Fig.~\ref{fig:example} presents a complex question and its corresponding program whose functions take KB elements (\textit{i.e.}, entities, relations and concepts) as arguments.
\textit{E.g.}, the relation $\mathtt{tourist\ attractions}$ is the argument of function \textit{Relate}.

For most KBs, the parallel question-program pairs are lacking because such annotation is both expensive and labor-intensive. 
Thus, the PI models have to learn only from question-answer pairs. Typically, they take the answers as weak supervision and search for gold programs with reinforcement learning (RL)~\cite{CPI,NSM,NPI}. 
The combinatorial explosion in program space, along with extremely sparse rewards, makes the learning challenging. 
Abundant attempts have been made to improve the stability of RL algorithms with pseudo-gold programs~\cite{NSM}, noise-stabilizing wrapper~\cite{NPI}, or auxiliary rewards~\cite{CPI}. 
Despite promising results, they require significant human efforts to develop carefully-designed heuristics or are constrained to relatively simple questions.

Recently, for several KBs, there emerge question-program annotation resources~\cite{clevr,KQAPro}. 
Thanks to the supervision signals (\textit{i.e.}, program annotation for each question), the PI models on these rich-resourced KBs achieve impressive performance for even extremely complex questions, and are free from expert engineering. 
Intuitively, leveraging these supervision signals to aid program induction for low-resourced KBs with only weak-supervision signals (\textit{i.e.}, question-answer pairs) is a promising direction.
In this paper, we formalize it as \textbf{Program Transfer}.

In practice, program transfer is challenging due to the following reasons:
(a) \textbf{Domain Heterogeneity}.  The questions and KBs across domains are both heterogeneous due to language and knowledge diversity~\cite{Lan2021ASO}. It is hard to decide what to transfer for program induction.
(b) \textbf{Unseen KB Elements.} The coverage of source KB is limited, \textit{e.g.}, KQA Pro in~\cite{KQAPro} covers only 3.9\% relations and 0.24\% concepts of Wikidata. Thus, most elements in the massive scale target KB are not covered in the source. 
(c) \textbf{Huge Search Space.} The search space of function arguments depends on the scale of target KB. For realistic KBs containing millions of entities, concepts and relations, the huge search space is unmanageable. 

To address the above problems, we propose a novel two-stage parsing framework with an efficient ontology-guided pruning strategy. 
First, we design a \textbf{sketch parser} to parse the question into a program sketch (the left side in Fig.~\ref{fig:example}), which is composed of functions without arguments. 
As~\citet{Baroni2019LinguisticGA} points out, the composition of functions well captures the language compositionality. Translation from questions to sketches is thus relevant to language compositional structure and independent of KB structure. Therefore, our sketch parser can transfer across KBs. 
Second, we design an \textbf{argument parser} to fill in the detailed arguments (typically KB elements) for functions in the sketch.
It retrieves relevant KB elements from the target KB and ranks them according to the question.
Specifically, it identifies KB elements with their label descriptions and relies on language understanding to resolve unseen ones. 
We further propose an \textbf{ontology-guided pruning} strategy, which introduces high-level KB ontology to prune the candidate space for the argument parser, thus alleviating the problem of huge search space. 

Specifically, the sketch parser is implemented with a Seq2Seq model with the attention mechanism.
The argument parser identifies elements through semantic matching and utilizes pre-trained language models \cite{bert} for language understanding.
The high-level ontology includes the domain and range of relations and entity types. 

In evaluation, we take the Wikidata-based KQA Pro as the source, Freebase-based ComplexWebQuestions and WebQuestionSP as the target domain datasets. 
Experimental results show that our method improves the F1 score by 14.7\% and 2.5\% respectively, compared with SOTA methods that learn from question-answer pairs.

\textbf{Our contributions} include: (a) proposing the approach of program transfer for Complex KBQA for the first time; (b) proposing a novel two-stage parsing framework with an efficient ontology-guided pruning strategy for program transfer; (c) demonstrating the effectiveness of program transfer through extensive experiments and careful ablation studies on two benchmark datasets.

\section{Related Work}

\noindent \textbf{KBQA}. KBQA aims to find answers for questions expressed in natural language from a KB, such as Freebase~\cite{freebase}, DBpedia~\cite{dbpedia} and Wikidata~\cite{Wikidata}. Current methods for KBQA can be categorized into two groups: 1) semantic parsing based methods~\cite{emnlp13, querygraph, NSM, NPI}, which learn a semantic parser that converts questions into intermediate logic forms which can be executed against a KB; 2) information retrieval based methods~\cite{emnlp14, acl14, kvmem, MetaQA, graphnet, pullnet,Shi2021TransferNetAE}, which retrieve candidate answers from the topic-entity-centric subgraph and then rank them according to the questions.
Recently, semantic parsing for KBQA has gained increasing research attention because the methods are effective and more interpretable.
Multiple kinds of logical forms have been proposed and researched, such as SPARQL~\cite{sparql}, $\lambda$-DCS~\cite{dcs}, $\lambda$-calculus~\cite{ccg}, query graph~\cite{querygraph}, program~\cite{NSM}. PI aims to convert questions into programs, and is in line with semantic parsing.

\noindent \textbf{Cross-domain Semantic Parsing.} Cross-domain semantic parsing trains a
semantic parser on some source domains and adapts it to the target domain.
Some works~\cite{Herzig2017NeuralSP, Su2017CrossdomainSP, Fan2017TransferLF}
pool together examples from multiple datasets in different domains and
train a single sequence-to-sequence model over
all examples, sharing parameters across domains. However, these methods rely on annotated logic forms in the target domain. To facilitate low-resource target domains, ~\cite{Chen2020LowResourceDA} adapts to
target domains with a very limited amount of annotated data. Other works consider a zero-shot semantic parsing task~\cite{Givoli2019ZeroShotSP}, decoupling structures from lexicons for transfer. However, they only learn from the source domain without further learning from the target domain using the transferred prior knowledge. In addition, existing works mainly focus on the domains in OVERNIGHT~\cite{Wang2015BuildingAS}, which are much smaller than large scale KBs such as Wikidata and Freebase. Considering the complex schema of large scale KBs, transfer in ours setting is more challenging.

\section{Problem Formulation}
\label{sec:preliminary}
In this section, we first give some necessary definitions and then formulate our task.

\noindent \textbf{Knowledge Bases}. Knowledge base describes concepts, entities, and the relations between them. It can
be formalized as $\mathcal{KB} = \{\mathcal{C}, \mathcal{E}, \mathcal{R}, \mathcal{T}\}$. $\mathcal{C}$,   $\mathcal{E}$, $\mathcal{R}$ and $\mathcal{T}$ denote the sets of concepts, entities, relations and triples respectively.
Relation set $\mathcal{R}$ can be formalized as $\mathcal{R} = \{r_e, r_c\} \cup \mathcal{R}_l$, where $r_e$ is $\mathtt{instanceOf}$, $r_c$ is $\mathtt{subClassOf}$, and $\mathcal{R}_l$ is the general relation set. $\mathcal{T}$ can be divided into three disjoint subsets: (1) $\mathtt{instanceOf}$ triple set $\mathcal{T}_e = \{(e, r_e, c) | e \in \mathcal{E}, c \in \mathcal{C}\}$; (2) $\mathtt{subClassOf}$ triple set $\mathcal{T}_c = \{(c_{i}, r_c, c_{j}) | c_{i}, c_{j} \in \mathcal{C}\}$; (3) relational triple set $\mathcal{T}_l = \{(e_i, r, e_j) | e_i, e_j \in \mathcal{E}, r \in \mathcal{R}_{l}\}$.

\noindent \textbf{Program}. Program is composed of symbolic functions with arguments, and produces an answer when executed against a KB. Each function defines a basic operation on KB and takes a specific type of argument. For example, the function \textit{Relate} aims to find entities that have a specific \textit{relation} with the
given entity. Formally, a program $y$ is denoted as $\left\langle o_1[arg_1], \cdots, o_t[arg_t], \cdots, o_{|y|}[arg_{|y|}] \right\rangle, o_t \in \mathcal{O}, arg_t \in \mathcal{E}\cup\mathcal{C}\cup\mathcal{R}$. Here, $\mathcal{O}$ is a pre-defined function set, which covers basic reasoning operations over KBs~\cite{KQAPro}. According to the argument type, $\mathcal{O}$ can be devided into four disjoint subsets: $\mathcal{O} = \mathcal{O^{E}}\cup\mathcal{O^{C}}\cup\mathcal{O^{R}}\cup\mathcal{O}^\emptyset$, representing the functions whose argument type is entity, concept, relation and empty respectively. Table~\ref{tab:program} gives some examples of program functions.

\begin{table}[t]
  \centering
  \scalebox{0.85}{
    \small
    \setlength{\tabcolsep}{2pt}
    \begin{tabular}{ccccc}
    \toprule
        Function & \begin{tabular}{c} Argument \\ Type\end{tabular} & Argument & Description \\
        \midrule
        \textit{Find} & entity & $\mathtt{FC\ Barcelona}$ & \begin{tabular}{c}
        Find the specific \\ KB entity \end{tabular}  \\
        \midrule
        \textit{Relate} & relation & $\mathtt{arena\ stadium}$ & \begin{tabular}{c} Find the entities that \\ hold a specific relation \\ with the given entity\end{tabular} \\
        \midrule
        \textit{FilterConcept} & concept & $\mathtt{sports\ facility}$ & \begin{tabular}{c} Find the entities that \\ belong to a specific \\ concept\end{tabular} \\
        \midrule
        \textit{And} & - & - & \begin{tabular}{c} Return the intersection \\ of two entity sets \end{tabular} \\
    \bottomrule
    \end{tabular}
    }
    \caption{Function examples. - means empty.}
    \label{tab:program}
\end{table}

\noindent \textbf{Program Induction}. Given a $\mathcal{KB}$, and a complex natural language question $x = \left\langle w_1, w_2, \cdots, w_{|x|} \right\rangle$, it aims to produce a program $y$ that generates the right answer $z$ when executed against $\mathcal{KB}$. 

\noindent \textbf{Program Transfer}. In this task, we have access to the source domain data $S = \left\langle \mathcal{KB}^{S}, \mathcal{D}^{S}\right\rangle$, where $\mathcal{D}^{S}$ contains pairs of question and program $\{(x_i^{S}, y_i^{S})\}_{i=1}^{n^{S}}$; and target domain data $T = \left\langle \mathcal{KB}^{T}, \mathcal{D}^{T}\right\rangle$, where $\mathcal{D}^{T}$ contains pairs of question and answer $\{(x_i^{T}, z_i^{T})\}_{i=1}^{n^{T}}$. We aim at learning a PI model to translate a question $x$ for $\mathcal{KB}^{T}$ into program $y$, which produces the correct answer when executed on $\mathcal{KB}^{T}$.

\begin{figure*}[htbp]
\includegraphics[width=\linewidth]{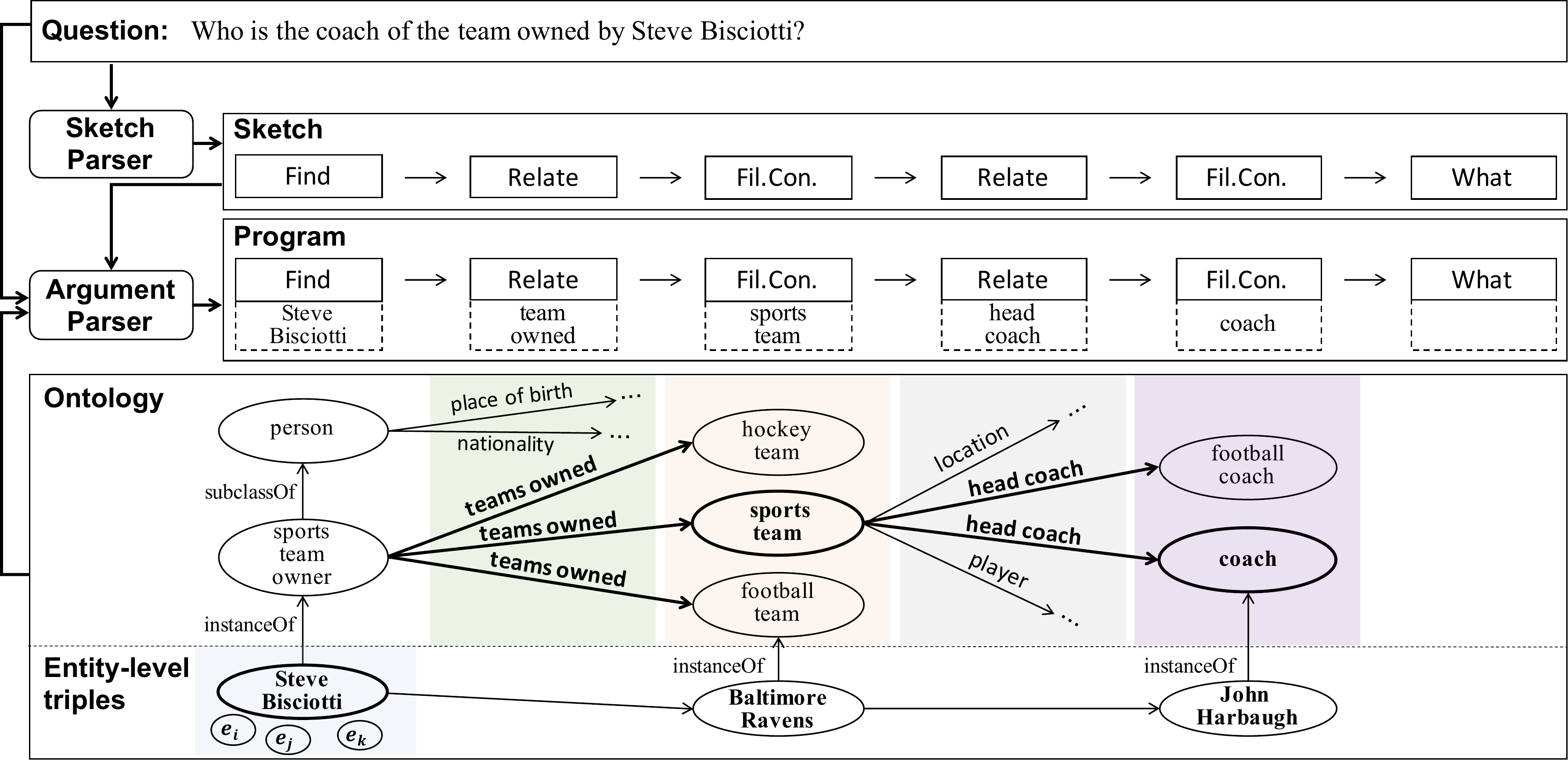}
\caption{We design a high-level sketch parser to generate the sketch, and a low-level argument parser to predict arguments for the sketch. The arguments are retrieved from candidate pools which are illustrated by the color blocks. The arguments for functions are mutually constrained by the ontology structure. For example, when the second function \textit{Relate} finds the argument $\mathtt{teams\ owned}$, the candidate pool for the third function \textit{Fil.Con.} (short for \textit{FilterConcept}) is reduced to the range of relation $\mathtt{teams\ owned}$. 
}
\label{fig:model}
\end{figure*}

\section{Framework}

As mentioned in the introduction, to perform program transfer for Complex KBQA, we need to address three crucial problems:
(1) What to transfer when both questions and KBs are heterogeneous? (2) How to deal with the KB elements unseen in the external annotations? (3) How to prune the search space of input arguments to alleviate the huge search space problem? In this section, we introduce our two-stage parsing framework with an ontology-guided pruning strategy, which is shown in Fig.~\ref{fig:model}.


(1) \textbf{Sketch Parser}: 
At the first stage, we design a sketch parser $f^{s}$ to parse $x$ into a program sketch $y_{s} = \left\langle o_1, \cdots, o_t, \cdots o_{|y|}\right\rangle$, which is a sequence of functions without arguments. The sketch parsing process can be formulated as 
\begin{equation}
    y_{s} = f^{s}(x).
\end{equation}
Translation from question to sketch is relevant to language compositionality, and irrelevant to KB structure. Therefore, the sketch parser can generalize across KBs.

(2) \textbf{Argument Parser}: At the second stage, we design an argument parser $f^{a}$ to retrieve the argument $arg_{t}$ from a candidate pool $\mathcal{P}$ for each function $o_{t}$, which can be formulated as 
\begin{equation}
    arg_{t} = f^{a}(x, o_t, \mathcal{P}).
\end{equation} 
Here, the candidate pool $\mathcal{P}$ contains the relevant elements in $\mathcal{KB}^{T}$, including concepts, entities, and relations. In a real KB, the candidate pool is usually huge, which makes searching and learning from answers very hard.
Therefore, we propose an ontology-guided pruning strategy, which dynamically updates the candidate pool and progressively reduces its search space.

In the following we will introduce the implementation details of our sketch parser (Section~\ref{sec:sketch parser}), argument parser (Section~\ref{sec:argument parser}) and training strategies (Section~\ref{sec:training}).

\subsection{Sketch Parser}
\label{sec:sketch parser}

The sketch parser is based on encoder-decoder model~\cite{Seq2Seq} with attention mechanism~\cite{seq2tree}. 
We aim to estimate $p(y_s|x)$, the conditional probability of sketch $y_s$ given input $x$. It can be decomposed as:
\begin{equation}
    p(y_s|x) = \prod_{t=1}^{|y_s|}p(o_t|o_{<t}, x),
\end{equation}
where $o_{<t} = o_1,...,o_{t-1}$.

Specifically, our sketch parser comprises a question encoder that encodes the question into vectors and a sketch decoder that autoregressively outputs the sketch step-by-step. The details are as follows:

\noindent \textbf{Question Encoder.}
We utilize BERT~\cite{bert} as the encoder. Formally,
\begin{equation}
\label{eq:bert}
\mathbf{\bar{x}}, (\mathbf{x}_1, \cdots, \mathbf{x}_i, \cdots, \mathbf{x}_{|x|}) = \text{BERT}(x),
\end{equation}
  where $\mathbf{\bar{x}} \in \mathbbm{R}^{\hat{d}}$ is the question embedding, and $\mathbf{x}_i \in \mathbbm{R}^{\hat{d}}$ is the hidden vector of word $x_i$. $\hat{d}$ is the hidden dimension.

\noindent \textbf{Sketch Decoder.}
We use Gated Recurrent Unit (GRU)~\cite{gru1}, a well-known variant of RNNs, as our decoder of program sketch. 
The decoding is conducted step by step. After we have predicted $o_{t-1}$, the hidden state of step $t$ is computed as:
\begin{equation}
    \label{eq:h_t}
    \mathbf{h}_t = \text{GRU}(\mathbf{h}_{t-1}, \mathbf{o}_{t-1}),
\end{equation}
where $\mathbf{h}_{t-1}$ is the hidden state from last time step, $\mathbf{o}_{t-1} = [\mathbf{W}]_{o_{t-1}}$ denotes the embedding corresponding to $o_{t-1}$ in the embedding matrix $\mathbf{W} \in \mathbbm{R}^{|\mathcal{O}|\times d}$. 
We use $\mathbf{h}_t$ as the attention key to compute scores for each word in the question based on the hidden vector $\mathbf{x}_i$, and compute the attention vector $\mathbf{c}_t$ as:
\begin{equation}
    \begin{aligned}
         \alpha_i &= \frac{\exp(\mathbf{x}^\mathrm{T}_i \mathbf{h}_t)}{\sum_{j=1}^{|x|} \exp(\mathbf{x}^\mathrm{T}_j \mathbf{h}_t)},\\ \mathbf{c}_t &=\sum_{i=1}^{|x|} \alpha_i \mathbf{x}_i.
    \end{aligned}
\end{equation}
The information of $\mathbf{h}_t$ and $\mathbf{c}_t$ are fused to predict the final probability of the next sketch token:
\begin{equation}\label{eq:gt}
    \begin{aligned}
         \mathbf{g}_t &= \mathbf{h}_t + \mathbf{c}_t, \\
         p(o_t|o_{<t}, x) &= \left[\text{Softmax}(\text{MLP}(\textbf{g}_t))\right]_{o_t},
    \end{aligned}
\end{equation}
where MLP (short for multi-layer perceptron) projects $\hat{d}$-dimensional feature to $|\mathcal{O}|$-dimension, which consists of two linear layers with ReLU activation.

\subsection{Argument Parser}
\label{sec:argument parser}

In the above section, the sketch is obtained with a sketch parser. In this section, we will introduce our argument parser, which aims to retrieve the argument $arg_{t}$ from the target KB for each function $o_t$ in the sketch. To reduce the search space, it retrieves arguments from a restricted candidate pool $\mathcal{P}$, which is constructed with our ontology-guided pruning strategy. In the following, we will introduce the argument retrieval process and the candidate pool construction process.

\noindent \textbf{Argument Retrieval}.
Specifically, we take $\mathbf{g}_t$ in Equation~\ref{eq:gt} as the context representation of $o_t$, learn vector representation $\mathbf{P}_{i} \in \mathbbm{R}^{\hat{d}}$ for each candidate $\mathcal{P}_{i}$, and calculate the probability for $\mathcal{P}_{i}$ based on $\mathbf{g}_t$ and $\mathbf{P}_{i}$. Candidate $\mathcal{P}_{i}$ is encoded with the BERT encoder in Equation~\ref{eq:bert}, which can be formulated as:
\begin{equation}
    \mathbf{P}_i = \text{BERT}(\mathcal{P}_i).
\end{equation}
$\mathbf{P}_{i}$ is the $i^{th}$ row of $\mathbf{P}$.
The probability of candidate $arg_t$ is calculated as:
\begin{equation}
    p(arg_t|x, o_t, \mathcal{P}) = [\text{Softmax}(\mathbf{P}\mathbf{g}_t)]_{arg_t}.
\end{equation}



\noindent \textbf{Candidate Pool Construction}.
In the following, we will introduce the KB ontology first. Then, we will describe the rationale of our ontology-guided pruning strategy and its implementation details.

In KB, The domain and range of relations, and the type of entities form the KB ontology. Specifically, a relation $r$ comes with a domain $dom(r) \subseteq \mathcal{C}$ and a range $ran(r) \subseteq \mathcal{C}$. An entity $e$ comes with a type $type(e) = \{ c | (e, \mathtt{instanceOf}, c) \in \mathcal{T} \}$.
For example, as shown in Fig.~\ref{fig:model},
$\mathtt{sports\ team\ owner} \in dom(\mathtt{teams\ owned})$, $\mathtt{sports\ team} \in ran(\mathtt{teams\ owned})$, and $\mathtt{sports\ team} \in type(\mathtt{Baltimore\ Ravens})$.

The rationale of our pruning is that the arguments for program functions are mutually constrained according to the KB ontology. Therefore, when the argument $arg_t$ for $o_t$ is determined, the possible candidates for $\{o_{i}\}_{i=t+1}^{|y_s|}$ will be adjusted. For example, in Fig.~\ref{fig:model}, when $\textit{Relate}$ takes $\mathtt{teams\ owned}$ as the argument, the candidate pool for the next \textit{FilterConcept} is constrained to the range of relation $\mathtt{teams\ owned}$, thus other concepts (\textit{e.g.}, $\mathtt{time\ zone}$) will be excluded from the candidate pool.

In practice, we propose a set of ontology-oriented operators to adjust the candidate pool $\mathcal{P}$ step-by-step. Specifically, we define
three ontology-oriented operators $C(e), R(r), D^{-}(c)$, which aim to find the type of entity $e$, the range of relation $r$, and the relations whose domain contains $c$. Furthermore, we use the operators to maintain an entity pool $\mathcal{P^{E}}$, a relation pool $\mathcal{P^{R}}$ and a concept pool $\mathcal{P^{C}}$. 
When $arg_t$ of $o_t$ is determined, we will update $\mathcal{P^{E}}$, $\mathcal{P^{R}}$, and $\mathcal{P^{C}}$ using $C(e), R(r), D^{-}(c)$. We take one of the three pools as $\mathcal{P}$ according to the argument type of $o_t$.  The detailed algorithm is shown in Appendix.

\subsection{Training}
\label{sec:training}
We train our model using the popular pretrain-finetune paradigm.
Specifically, we pretrain the parsers on the source domain data $\mathcal{D}^{S} = \left\{\left(x_i^{S}, y_i^{S}\right)\right\}_{i=1}^{n^{S}}$ in a supervised way.
After that, we conduct finetuning on the target domain data $\mathcal{D}^{T} = \left\{\left(x_i^{T}, z_i^{T}\right)\right\}_{i=1}^{n^{T}}$ in a weakly supervised way.


\noindent \textbf{Pretraining in Source Domain.}
Since the source domain data provides complete annotations, we can directly maximize the log-likelihood of the golden sketch and golden arguments:

\begin{equation}\label{eq:pretrain}
\begin{aligned}
     \mathcal{L}^\text{pretrain} &= -  \sum_{(x^{S}, y^{S}) \in \mathcal{D}^{S}}
     \bigg( \log p(y_s^{S} | x^{S}) \\ 
     &+\  \sum_{t=1}^{|y_s|}  \log p(arg_t^{S}|x^{S}, o_t^{S}, \mathcal{P}) \bigg).
\end{aligned}   
\end{equation}

\noindent \textbf{Finetuning in Target Domain.}
At this training phase, questions are labeled with answers while programs remain unknown. 
The basic idea is to search for potentially correct programs and optimize their corresponding probabilities.
Specifically, we propose two training strategies:
\begin{itemize}[leftmargin=13pt]
\setlength{\itemsep}{2pt}
\setlength{\parskip}{0pt}
\setlength{\parsep}{0pt}
    \item Hard-EM Approach. At each training step, hard-EM generates a set of possible programs with beam search based on current model parameters, and then executes them to find the one whose answers have the highest F1 score compared with the gold. Let $\hat{y}^{T}$ denote the best program, we directly maximize $p(\hat{y}^{T} | x^{T})$ like Equation~\ref{eq:pretrain}.
    
    \item Reinforcement learning (RL). It formulates the program generation as a decision making procedure and computes the rewards for sampled programs based on their execution results. We take the F1 score between the executed answers and golden answers as the reward value, and use REINFORCE~\cite{williams1992simple} algorithm to optimize the parsers.
\end{itemize}

\section{Experimental Settings}

\subsection{Datasets}
\noindent \textbf{Source Domain.}
KQA Pro~\cite{KQAPro} provides 117,970 question-program pairs based on a Wikidata~\cite{Wikidata} subset. 

\noindent \textbf{Target Domain.}
We use WebQuestionSP (WebQSP)~\cite{WebQSP} and ComplexWebQuestions (CWQ)~\cite{ComplexWebQ} as the target domain datasets for two reasons: (1) They are two widely used benchmark datasets in Complex KBQA; (2) They are based on a large-scale KB Freebase~\cite{freebase}, which makes program transfer challenging.
Specifically, WebQSP contains 4,737 questions and is divided into 2,998 train, 100 dev and 1,639 test cases. 
CWQ is an extended version of WebQSP which is more challenging, with four types of questions: composition (44.7\%), conjunction (43.6\%), comparative (6.2\%), and superlative (5.4\%).
CWQ is divided into 27,639 train, 3,519 dev and 3,531 test cases.
We use the Freebase dump on 2015-08-09\footnote{http://commondatastorage.googleapis.com/freebase-public/rdf/freebase-rdf-latest.gz}, from which we extract the type of entities, domain and range of relations to construct the ontology. The average domain, range, type size is 1.43 per relation, 1.17 per relation, 8.89 per entity respectively.

Table~\ref{tab:kb_statistics} shows the statistics of the source and target domain KB. The target domain KB contains much more KB elements, and most of them are uncovered by the source domain.

\begin{table}[htbp]
	\centering
	\small
		\begin{tabular}{cccc}
			\toprule
			Domain & \# Entities & \# Relations & \# Concepts \\
			\midrule
			Source 	& 16,960 & 363 & 794 \\
			Target & 30,943,204 &  15,015 & 2,519 \\
			\bottomrule
		\end{tabular}%
    \caption{The statistics for source and target domain KB.}
    \label{tab:kb_statistics}
\end{table}%

\subsection{Baselines}
In our experiments, we select representative models that learn from question-answer pairs as our baselines. They can be categorized into three groups: program induction methods, query graph generation methods and information retrieval methods.

Existing program induction methods search for gold programs with RL. They usually require human efforts or are constrained to simple questions. \textbf{NSM}~\cite{NSM} uses the provided entity, relation and type annotations to ease the search, and can solve relatively simple questions.
\textbf{NPI}~\cite{NPI} designs heuristic rules such as disallowing repeating or useless actions for efficient search.

Existing query graph generation methods generate query graphs whose execution on KBs produces the answer. They use entity-level triples as search guidance, ignoring the useful ontology.
\textbf{TEXTRAY}~\cite{TEXTRAY} uses a decompose-execute-join approach.
\textbf{QGG}~\cite{QGG} incorporates constraints into query graphs in the early stage.
\noindent \textbf{TeacherNet}~\cite{NSM-WSDM} utilizes bidirectional searching.

Existing information retrieval methods directly construct a question-specific sub-KB and then rank the entities in the sub-KB to get the answer. \textbf{GraftNet}~\cite{graphnet} uses heuristics to create the subgraph and uses a variant of graph convolutional networks to rank the entities. \textbf{PullNet}~\cite{pullnet} improves GraftNet by iteratively constructing the subgraph instead of using heuristics.

Besides, we compare our full model \textbf{$\text{Ours}$} with $\text{Ours}_{\text{-f}}$, $\text{Ours}_{\text{-p}}$, $\text{Ours}_{\text{-pa}}$, $\text{Ours}_{\text{-o}}$, which denotes our model without finetuning, without pretraining, without pretraining of argument parser, and without our ontology-guided pruning strategy respectively.







\subsection{Evaluation Metrics} 
Following prior works~\cite{emnlp13, graphnet, NSM-WSDM}, we use F1 score and Hit@1 as the evaluation metrics. Since questions in the datasets have multiple answers, F1 score reflects the coverage of predicted answers better.

\subsection{Implementations}

We used the bert-base-cased model of HuggingFace\footnote{https://github.com/huggingface/transformers} as our BERT encoder with the hidden dimension $\hat{d}$ 768. 
The hidden dimension of the sketch decoder $d$ was 1024. 
We used AdamW~\cite{AdamW} as our optimizer.
We searched the learning rate for BERT paramters in \{1e-4, 3e-5, 1e-5\}, the learning rate for other parameters in \{1e-3, 1e-4, 1e-5\}, and the weight decay in \{1e-4, 1e-5, 1e-6\}. 
According to the performance on dev set, we finally used learning rate 3e-5 for BERT parameters, 1e-3 for other parameters, and weight decay 1e-5. 






\section{Experimental Results}

\begin{table}[htbp]
	\centering%
	\small
		\begin{tabular}{ccccc}
		    \toprule	
		    \multirow{2}{*}{Models}&\multicolumn{2}{c}{WebQSP}&\multicolumn{2}{c}{CWQ}\\
			\cmidrule(lr){2-3} \cmidrule(lr){4-5}
			& F1 & Hit@1& F1 & Hit@1 \\
			\midrule
   NSM & - & \multicolumn{1}{c}{69.0} & - & - \\
   NPI & - & \multicolumn{1}{c}{72.6} & - & - \\

   TEXTRAY & 60.3 & \multicolumn{1}{c}{72.2} & 33.9 & 40.8 \\
   QGG & \underline{74.0} & \multicolumn{1}{c}{-} & 40.4 & 44.1 \\
  TeacherNet & 67.4 & \multicolumn{1}{c}{\underline{74.3}} & \underline{44.0} & \underline{48.8} \\
     GraftNet & 62.3 & \multicolumn{1}{c}{68.7} & - & 32.8* \\
   PullNet & - & \multicolumn{1}{c}{68.1} & - & 47.2* \\
   



   \midrule
			$\text{Ours}_{\text{-f}}$	&	53.8 & \multicolumn{1}{c}{53.0} & 45.9 & 45.2 \\
			$\text{Ours}_{\text{-p}}$	&	3.2 & \multicolumn{1}{c}{3.1} & 2.3 &	2.1 \\
			$\text{Ours}_{\text{-pa}}$ & 70.8 & \multicolumn{1}{c}{68.9} & 54.5 & 54.3 \\
			$\text{Ours}_{\text{-o}}$ & 72.0 & \multicolumn{1}{c}{71.3} & 55.8 & 54.7\\
   \textbf{\text{Ours}} & \textbf{76.5} & \multicolumn{1}{c}{\textbf{74.6}} & \textbf{58.7} & \textbf{58.1} \\
			\bottomrule
	\end{tabular}%
	\caption{Performance comparison of different methods (F1 score and Hits@1 in percent). We highlight the best results in bold and second with an underline. *: reported by PullNet on the dev set.}
	\label{tab:exp}
\end{table}%

\subsection{Overall Results}

As shown in Table~\ref{tab:exp}, our model achieves the best performance on both WebQSP and CWQ.
Especially on CWQ, we have an absolute gain of 14.7\% in F1 and 9.3\% in Hit@1, beating previous methods by a large margin. 
Note that CWQ is much more challenging than WebQSP because it includes more compositional and conjunctional questions.
Previous works mainly suffer from the huge search space and sparse training signals. 
We alleviate these issues by transferring the prior knowledge from external annotations and incorporating the ontology guidance. Both of them reduce the search space substantially.
On WebQSP, we achieve an absolute gain of 2.5\% and 0.3\% in F1 and Hit@1, respectively, demonstrating that our model can also handle simple questions well, and can adapt to different complexities of questions.

Note that our F1 scores are higher than the corresponding Hit@1.
This is because we just randomly sampled one answer from the returned answer set as the top 1 without ranking them.


\begin{table}[htbp]
	\centering
	\small
		\begin{tabular}{ccc}
			\toprule
			Models & WebQSP & CWQ \\
			\midrule
			Top-1	& 76.5 & 58.7 \\
			Top-2	& 81.1 & 61.2 \\
			Top-5	& 85.4 & 63.3\\
			Top-10  & 86.9 & 65.0\\
			\bottomrule
		\end{tabular}%
    \caption{The highest F1 score in the top-k programs.}
	\label{tab:topk}
\end{table}%

We utilize beam search to generate multiple possible programs and evaluate their performance. 
Table~\ref{tab:topk} shows the highest F1 score in the top-k generated programs, where top-1 is the same as Table~\ref{tab:exp}. 
We can see that the best F1 in the top-10 programs is much higher than the F1 of the top-1 (\textit{e.g.}, with an absolute gain 10.4\% for WebQSP and 6.3\% for CWQ). 
This indicates that a good re-ranking method can further improve the overall performance of our model. We leave this as our future work.

\subsection{Ablation study}

\noindent \textbf{Pretraining: }
As shown in Table~\ref{tab:exp}, when comparing $\text{Ours}_{\text{-pa}}$ with $\text{Ours}$, the F1 and Hit@1 on CWQ drop by $4.2\%$ and $3.8\%$ respectively, which indicates that the pretraining for the argument parser is necessary. 
$\text{Ours}_{\text{-p}}$ denotes the model without pretraining for neither sketch parser nor argument parser. 
We can see that its results are very poor, achieving just about 3\% and 2\% on WebQSP and CWQ, indicating that the pretraining is essential, especially for the sketch parser. 

\noindent \textbf{Finetuning: }
Without finetuning on the target data, \textit{i.e.}, in $\text{Ours}_{\text{-f}}$, performance drops a lot compared with the complete model.
For example, F1 and Hit@1 on CWQ drop by 12.8\% and 12.9\% respectively.
It indicates that finetuning is necessary for the model's performance. As shown in Table~\ref{tab:kb_statistics}, most of the relations and concepts in the target domain are uncovered by the source domain.
Due to the semantic gap between source and target data, the prior knowledge must be properly transferred to the target domain to bring into full play.

\noindent \textbf{Ontology: }
We implemented $\text{Ours}_{\text{-o}}$ by removing ontology from KB and removing \textit{FilterConcept} from the program. 
Comparing $\text{Ours}_{\text{-o}}$ with $\text{Ours}$, the F1 and Hit@1 on CWQ drops by 2.9\% and 3.4\% respectively, which demonstrates the importance of ontology-guided pruning strategy. 
We calculated the search space size for each compositional and conjunctive question in the dev set of CWQ, and report the average size in Table~\ref{tab:searchspace}. The statistics shows that, the average search space size of $\text{Ours}$ is only 0.26\% and 3.2\% of that in $\text{Ours}_{\text{-o}}$ for the two kinds of questions. By incorporating the ontology guidance, $\text{Ours}$ substantially reduces the search space.



\begin{table}[htbp]
	\centering
	\small
		\begin{tabular}{ccc}
			\toprule
			Model & Composition & Conjunction \\
			\midrule
			$\text{Ours}_{\text{-o}}$	& 4,248,824.5 & 33,152.1 \\
			$\text{Ours}$ & 11,200.7 &  1,066.5 \\
			\bottomrule
		\end{tabular}%
    \caption{The average search space size for composition and conjunction questions in CWQ  set for $\text{Ours}$ and $\text{Ours}_{\text{-o}}$. }
    \label{tab:searchspace}
\end{table}%

\noindent \textbf{Hard-EM \textit{v.s.} RL:}
For both WebQSP and CWQ, training with Hard-EM achieves better performance. For RL, we simply employed the REINFORCE algorithm
and did not implement any auxiliary reward strategy since this is not the focus of our work. 
The sparse, delayed reward causes high variance, instability, and local
minima issues, making the training hard~\cite{CPI}.  
We leave exploring more complex training strategies as our future work.

\begin{table}[htbp]
	\centering
	\small
		\begin{tabular}{ccccccc}
			\toprule
			\multirow{2}{*}{Models}&\multicolumn{2}{c}{WebQSP}&\multicolumn{2}{c}{CWQ}\\
			\cmidrule(lr){2-3} \cmidrule(lr){4-5}
			& F1 & Hit@1& F1 & Hit@1 \\
			\midrule
			Hard-EM	&	76.5 & \multicolumn{1}{c}{74.6} & 58.7 & 58.1 \\
			RL	&	71.4 & \multicolumn{1}{c}{72.0} & 46.1 & 45.4 \\
			\bottomrule
		\end{tabular}%
	\caption{Results of different training strategies.}
	\label{tab:training}
\end{table}%

\subsection{Case Study}
\begin{figure}[htbp]
\includegraphics[width=\linewidth]{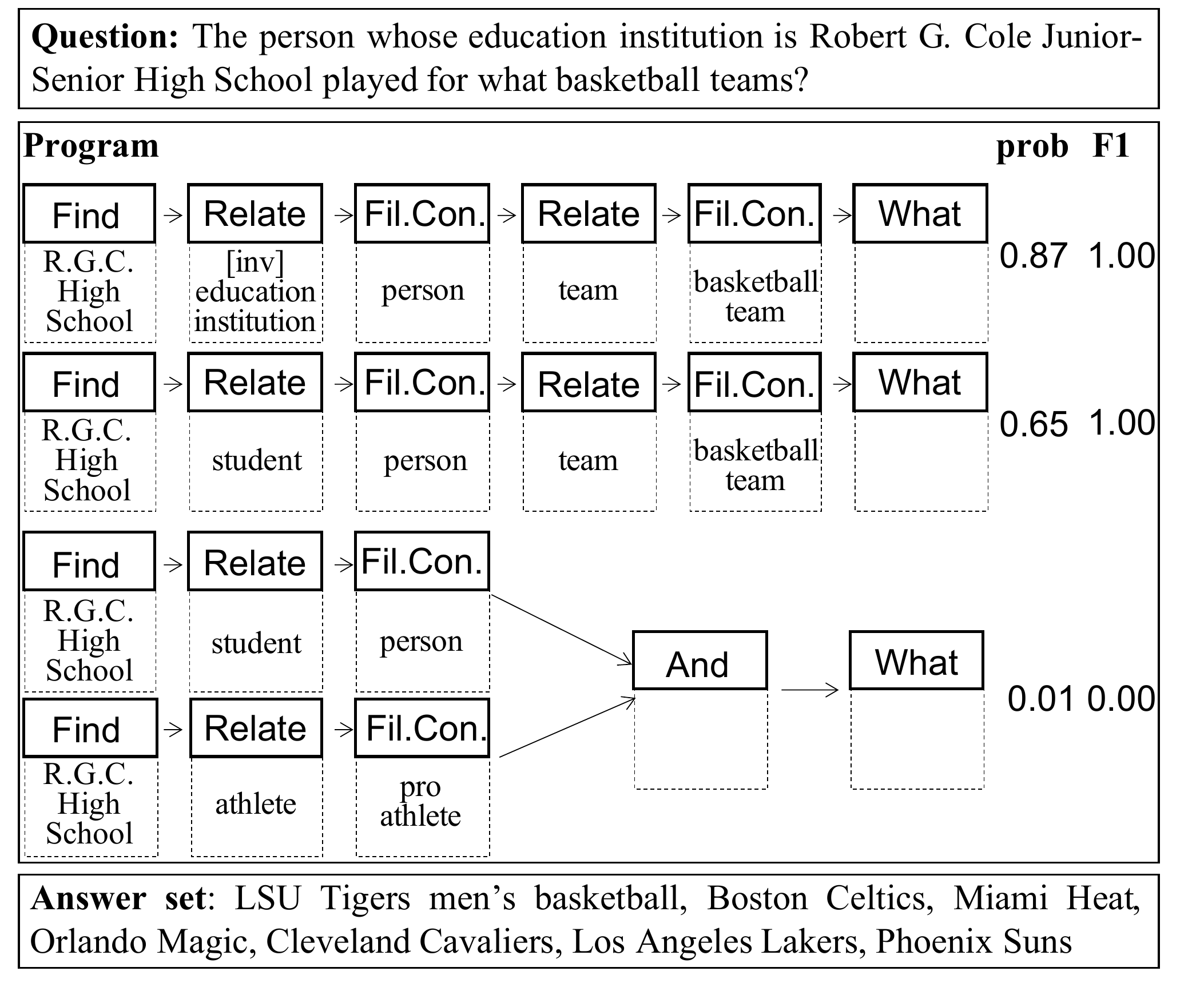}
\caption{An example from CWQ dev set. Our model translates the question into multiple programs with the corresponding probability and F1 score. We show the best, 2-nd best and 10-th best programs. Both the best and 2-nd best programs are correct.}
\label{fig:case}
\end{figure}
Fig.~\ref{fig:case} gives a case, where our model parses an question into multiple programs along with their probablility scores and F1 scores of executed answers.
Given the question ``\textit{The person whose education institution is Robert G. Cole Junior-Senior High School played for what basketball teams?}'', we show the programs with the largest, 2-nd largest and 10-th largest possibility score.
Both of the top-2 programs get the correct answer set and are semantically equivelant with the question, while the 10-th best program is wrong.  
\label{sec:casestudy}

			
			

\noindent \textbf{Error Analysis}
We randomly sampled 100 error cases whose F1 score is lower than 0.1 for manual inspection. The errors can be summarized into the following categories: (1) Wrong relation (53\%): wrongly predicted relation makes the program wrong, \textit{e.g.}, for question ``\textit{      What language do people in the Central Western Time Zone speak?}'', our model predicts the relation $\mathtt{main\ country}$, while the ground truth is $\mathtt{countries\ spoken\ in}$;
(2) Wrong concept (38\%): wrongly predicted concept makes the program wrong, \textit{e.g.}, for the question ``\textit{What continent does the leader Ovadia Yosel live in?}'', our model predicted the concept $\mathtt{location}$, whereas the ground truth is \textit{continent}.
(3) Model limitation (9\%): Handling attribute constraint was not considered in our model, \textit{e.g.}, for the question ``\textit{Who held his governmental position from before April 4, 1861 and influenced Whitman's poetry?}'', the time constraint \textit{April 4, 1861} cannot be handled.
			
			
			

\section{Conclusion}
In this parper, we propose program transfer for Complex KBQA for the first time.
We propose a novel two-stage parsing framework with an efficient ontology-guided pruning strategy. 
First, a sketch parser translates a question into the program, and then an argument parser fills in the detailed arguments for functions, whose search space is restricted by an ontology-guided pruning strategy.
The experimental results demonstrate that our program transfer approach outperforms the previous methods significantly. The ablation studies show that our two-stage parsing paradigm and ontology-guided pruning are both effective.

\section{Acknowledgments}

This work is supported by the National Key Research and Development Program of China (2020AAA0106501), the NSFC Youth Project (62006136), the Key-Area Research and Development Program of Guangdong Province (2019B010153002), grants from the Institute for Guo Qiang, Tsinghua University (2019GQB0003), Beijing Academy of Artificial Intelligence, and Huawei Noah’s Ark Lab.



\bibliography{anthology,custom}
\bibliographystyle{acl_natbib}

\clearpage

\appendix



\section{Ontology-guided Pruning}
\begin{algorithm}[H]
\caption{Ontology-guided Pruning\label{alg:construct}}
\textbf{Input:} natural language question $x$, program sketch $y_s$, knowledge base $\mathcal{KB} = \{\mathcal{C}, \mathcal{E}, \mathcal{R}, \mathcal{T}\}$\\
\textbf{Output:} $\{arg_t\}_{t=1}^{|y_s|}$
    \begin{algorithmic}
    \State $\mathcal{P^{E}} \gets \mathcal{E}, \mathcal{P^{R}} \gets \mathcal{R}, \mathcal{P^{C} \gets \mathcal{C}}, \mathcal{P} \gets \emptyset$
    \ForAll {$o_t$ in $y_s$}
        \If{$o_t \in \mathcal{O^{E}}$}
            \State $\mathcal{P} \gets \mathcal{P^{E}}$
            \State $arg_t = f^a(x, o_t, \mathcal{P})$
            \State $\mathcal{P^{C}} \gets C(arg_t)$
            \State $\mathcal{P^{R}} \gets \bigcup\limits_{c \in \mathcal{P^{C}}}{D^{-}(c)}$
        \ElsIf{$o_t \in \mathcal{O^{R}}$}
            \State $\mathcal{P} \gets \mathcal{P^{R}}$
            \State $arg_t = f^a(x, o_t, \mathcal{P})$
            \State $\mathcal{P^{C}} \gets R(arg_t)$
        \ElsIf{$o_t \in \mathcal{O^{C}}$}
            \State $\mathcal{P} \gets \mathcal{P^{C}}$
            \State $arg_t = f^a(x, o_t, \mathcal{P})$
            \State $\mathcal{P^{R}} \gets D^{-}(arg_t)$
        \EndIf
    \EndFor

\end{algorithmic}
\end{algorithm}



\section{Freebase Details}
We extracted a subset of Freebase which
contains all facts that are within 4-hops of
entities mentioned in the questions of CWQ and WebQSP. We extracted the domain constraint for relations according to `` \textit{/type/property/schema}'', range constraint for relations according to ``\textit{/type/property/expected\_type}'', type constraint for entities according to
``\textit{/type/type/instance}''. CVT nodes in the Freebase were dealed with concatenation of neiborhood relations.

\section{Program}
We list the functions of KQA Pro in Table~\ref{tab:functions}. The arguments in our paper are the textual inputs. To reduce the burden of the argument parser, for the functions that take multiple textual inputs, we concatenate them to a single input. 
\begin{table*}[htbp]
\centering
\scriptsize

\begin{tabularx}{\textwidth}{|l *{3}{Y}|}
\toprule
\textbf{Function} & \textbf{Functional Inputs $\times$ Textual Inputs $\rightarrow$ Outputs} & \textbf{Description} & \textbf{Example} (only show textual inputs) \\
\midrule

\textit{FindAll} &   () $\times$ () $\rightarrow$ (\textit{Entities})   &  Return all entities in KB  & - \\
\hline

\textit{Find} & () $\times$ (\textit{Name}) $\rightarrow$ (\textit{Entities})  &  Return all entities with the given name  &  \textit{Find(Kobe Bryant)} \\
\hline

\textit{FilterConcept} & (\textit{Entities}) $\times$ (\textit{Name}) $\rightarrow$ (\textit{Entities})  &  Find those belonging to the given concept  &  \textit{FilterConcept(athlete)} \\
\hline

\textit{FilterStr} & (\textit{Entities}) $\times$ (\textit{Key}, \textit{Value}) $\rightarrow$ (\textit{Entities}, \textit{Facts}) & Filter entities with an attribute condition of string type, return entities and corresponding facts  &  \textit{FilterStr(gender, male)} \\
\hline

\textit{FilterNum} & (\textit{Entities}) $\times$ (\textit{Key}, \textit{Value}, \textit{Op}) $\rightarrow$ (\textit{Entities}, \textit{Facts}) & Similar to \textit{FilterStr}, except that the attribute type is number  &  \textit{FilterNum(height, 200 centimetres, $>$)} \\
\hline

\textit{FilterYear} & (\textit{Entities}) $\times$ (\textit{Key}, \textit{Value}, \textit{Op}) $\rightarrow$ (\textit{Entities}, \textit{Facts}) & Similar to \textit{FilterStr}, except that the attribute type is year  &  \textit{FilterYear(birthday, 1980, $=$)} \\
\hline

\textit{FilterDate} & (\textit{Entities}) $\times$ (\textit{Key}, \textit{Value}, \textit{Op}) $\rightarrow$ (\textit{Entities}, \textit{Facts}) & Similar to \textit{FilterStr}, except that the attribute type is date  &  \textit{FilterDate(birthday, 1980-06-01, $<$)} \\
\hline

\textit{QFilterStr} & (\textit{Entities}, \textit{Facts}) $\times$ (\textit{QKey}, \textit{QValue}) $\rightarrow$ (\textit{Entities}, \textit{Facts}) & Filter entities and corresponding facts with a qualifier condition of string type  &  \textit{QFilterStr(language, English)} \\
\hline

\textit{QFilterNum} & (\textit{Entities}, \textit{Facts}) $\times$ (\textit{QKey}, \textit{QValue}, \textit{Op}) $\rightarrow$ (\textit{Entities}, \textit{Facts}) & Similar to \textit{QFilterStr}, except that the qualifier type is number  &  \textit{QFilterNum(bonus, 20000 dollars, $>$)} \\
\hline

\textit{QFilterYear} & (\textit{Entities}, \textit{Facts}) $\times$ (\textit{QKey}, \textit{QValue}, \textit{Op}) $\rightarrow$ (\textit{Entities}, \textit{Facts}) & Similar to \textit{QFilterStr}, except that the qualifier type is year  &  \textit{QFilterYear(start time, 1980, $=$)} \\
\hline

\textit{QFilterDate} & (\textit{Entities}, \textit{Facts}) $\times$ (\textit{QKey}, \textit{QValue}, \textit{Op}) $\rightarrow$ (\textit{Entities}, \textit{Facts}) & Similar to \textit{QFilterStr}, except that the qualifier type is date  &  \textit{QFilterDate(start time, 1980-06-01, $<$)} \\
\hline

\textit{Relate} & (\textit{Entity}) $\times$ (\textit{Pred}, \textit{Dir})
$\rightarrow$ (\textit{Entities}, \textit{Facts}) & Find entities that have a specific relation with the given entity & \textit{Relate(capital, forward)} \\
\hline

\textit{And} & (\textit{Entities}, \textit{Entities}) $\times$ () $\rightarrow$ (\textit{Entities}) & Return the intersection of two entity sets & - \\
\hline
\textit{Or} & (\textit{Entities}, \textit{Entities}) $\times$ () $\rightarrow$ (\textit{Entities}) & Return the union of two entity sets & - \\
\hline

\textit{QueryName} & (\textit{Entity}) $\times$ () $\rightarrow$ (string) & Return the entity name & - \\
\hline

\textit{Count} & (\textit{Entities}) $\times$ () $\rightarrow$ (number) & Return the number of entities & - \\
\hline

\textit{QueryAttr} & (\textit{Entity}) $\times$ (\textit{Key}) $\rightarrow$ (\textit{Value}) & Return the attribute value of the entity & \textit{QueryAttr(height)} \\
\hline

\textit{QueryAttrUnderCondition} & (\textit{Entity}) $\times$ (\textit{Key}, \textit{QKey}, \textit{QValue}) $\rightarrow$ (\textit{Value}) & Return the attribute value, whose corresponding fact should satisfy the qualifier condition & \textit{QueryAttrUnderCondition(population, point in time, 2016)} \\
\hline

\textit{QueryRelation} & (\textit{Entity}, \textit{Entity}) $\times$ ()
$\rightarrow$ (\textit{Pred}) & Return the predicate between two entities & \textit{QueryRelation(Kobe Bryant, America)} \\
\hline

\textit{SelectBetween} & (\textit{Entity}, \textit{Entity}) $\times$ (\textit{Key}, \textit{Op}) $\rightarrow$ (string) & From the two entities, find the one whose attribute value is greater or less and return its name & \textit{SelectBetween(height, greater)} \\
\hline

\textit{SelectAmong} & (\textit{Entities}) $\times$ (\textit{Key}, \textit{Op}) $\rightarrow$ (string) & From the entity set, find the one whose attribute value is the largest or smallest & \textit{SelectAmong(height, largest)} \\
\hline

\textit{VerifyStr} & (\textit{Value}) $\times$ (\textit{Value}) $\rightarrow$ (boolean) & Return whether the output of \textit{QueryAttr} or \textit{QueryAttrUnderCondition} and the given value are equal as string & \textit{VerifyStr(male)} \\
\hline

\textit{VerifyNum} & (\textit{Value}) $\times$ (\textit{Value}, \textit{Op}) $\rightarrow$ (boolean) & Return whether the two numbers satisfy the condition & \textit{VerifyNum(20000 dollars, $>$)} \\
\hline

\textit{VerifyYear} & (\textit{Value}) $\times$ (\textit{Value}, \textit{Op}) $\rightarrow$ (boolean) & Return whether the two years satisfy the condition & \textit{VerifyYear(1980, $>$)} \\
\hline

\textit{VerifyDate} & (\textit{Value}) $\times$ (\textit{Value}, \textit{Op}) $\rightarrow$ (boolean) & Return whether the two dates satisfy the condition & \textit{VerifyDate(1980-06-01, $>$)} \\
\hline

\textit{QueryAttrQualifier} & (\textit{Entity}) $\times$ (\textit{Key}, \textit{Value}, \textit{QKey})
$\rightarrow$ (\textit{QValue}) & Return the qualifier value of the fact (\textit{Entity}, \textit{Key}, \textit{Value}) & \textit{QueryAttrQualifier(population, 23,390,000, point in time)} \\
\hline

\textit{QueryRelationQualifier} & (\textit{Entity}, \textit{Entity}) $\times$ (\textit{Pred}, \textit{QKey})
$\rightarrow$ (\textit{QValue}) & Return the qualifier value of the fact (\textit{Entity}, \textit{Pred}, \textit{Entity}) & \textit{QueryRelationQualifier(spouse, start time)} \\

\bottomrule
\end{tabularx}
\caption{\label{tab:functions}Details of 27 functions in KQA Pro. Each function has 2 kinds of inputs: the functional inputs come from the output of previous functions, while the textual inputs come from the question.}
\end{table*}

\end{document}